%% file: bare_jrnl.tex
\documentclass[journal]{IEEEtran}
\usepackage{times}
\usepackage{epsfig}
\usepackage{graphicx}
\usepackage{amsmath}
\usepackage{amssymb}
\usepackage{multirow}
\usepackage{caption}
\usepackage{color}
\usepackage{booktabs}
\usepackage{nth}
\usepackage{float}
\usepackage{bbding}
\usepackage{xspace}
\usepackage{tabularx}
\usepackage{setspace}
\usepackage{bbm}
\usepackage{subcaption}
\usepackage{stackengine}
\usepackage{placeins}
\usepackage{footnote}
\usepackage{cite}
\usepackage[dvipsnames]{xcolor}
\newcommand{\vphan}{\vphantom{$\int^1_0$}}
\newcommand{\ie}{{\emph{i.e.}}\xspace}
\newcommand{\eg}{{\emph{e.g.}}\xspace}
\newcommand{\Eg}{{\emph{E.g.}}\xspace}
\newcommand{\vs}{{\emph{v.s.}}\xspace}
\newcommand{\etal}{{\emph{et al.}}\xspace}

\begin{document}

\title{Instance-Specific Feature Propagation for Referring Segmentation}

\author{Chang~Liu,
        Xudong~Jiang,~\IEEEmembership{Fellow,~IEEE},~
        Henghui~Ding
\thanks{Chang Liu, Xudong Jiang, Henghui Ding are with School of Electrical and Electronic
Engineering (EEE), Nanyang Technological University, Singapore 639798
(e-mail: liuc0058@e.ntu.edu.sg; exdjiang@ntu.edu.sg; ding0093@ntu.edu.sg) (Corresponding author: Henghui Ding.)}
\thanks{Manuscript received June 25, 2021.}}

\markboth{IEEE TRANSACTIONS ON MULTIMEDIA}%
{Liu \MakeLowercase{\textit{et al.}}: Instance-Specific Feature Propagation for Referring Segmentation}

\maketitle

\begin{abstract}
Referring segmentation aims to generate a segmentation mask for the target instance indicated by a natural language expression. There are typically two kinds of existing methods: one-stage methods that directly perform segmentation on the fused vision and language features; and two-stage methods that first utilize an instance segmentation model for instance proposal and then select one of these instances via matching them with language features. In this work, we propose a novel framework that simultaneously detects the target-of-interest via feature propagation and generates a fine-grained segmentation mask. In our framework, each instance is represented by an Instance-Specific Feature (ISF), and the target-of-referring is identified by exchanging information among all ISFs using our proposed Feature Propagation Module (FPM). Our instance-aware approach learns the relationship among all objects, which helps to better locate the target-of-interest than one-stage methods. Comparing to two-stage methods, our approach collaboratively and interactively utilizes both vision and language information for synchronous identification and segmentation. In the experimental tests, our method outperforms previous state-of-the-art methods on all three RefCOCO series datasets.
\end{abstract}

\begin{IEEEkeywords}
Referring segmentation, instance-specific, feature propagation.
\end{IEEEkeywords}

\IEEEpeerreviewmaketitle

\section{Introduction}

\begin{figure}[t]
   \begin{center}
      \includegraphics[width=\linewidth]{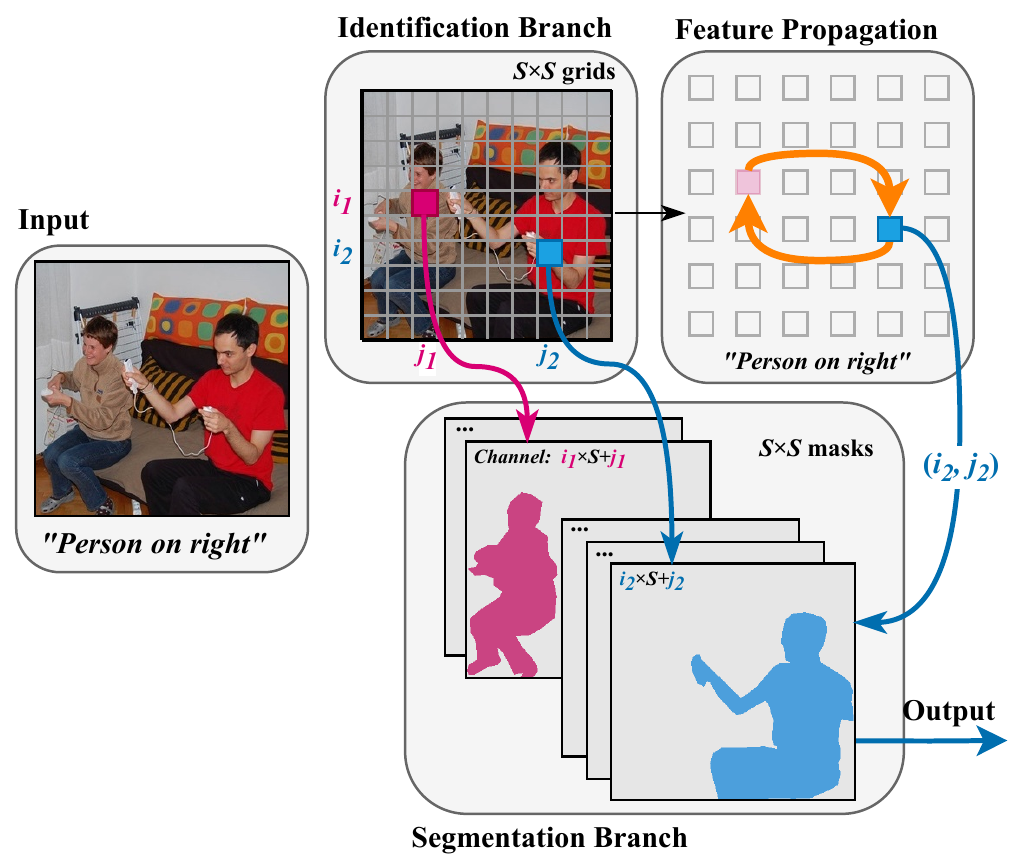}
   \end{center}
   \caption[]{Illustration of our method. We spatially divide the image into grid and assign each instance to a grid box where its center falls in. The model learns an Instance-Specific Feature for each grid as the representation of the corresponding instance, identifies the target and generates the mask simultaneously.\footnotemark}
   \label{fig:fig1}
\end{figure}

\IEEEPARstart{A}{s} deep learning methods are showing excellent performance in multiple areas including computer vision~\cite{he2016deep,ding2020semantic,liu2021towards,ding2021interaction,wang2021discovering,ding2022deep,shuai2018toward,liu2019feature}, natural language~\cite{chung2014empirical}, video processing~\cite{hu2020dipnet}, and interdisciplinary information~\cite{hu2020uncertainty,zhang2021prototypical,kuen2019scaling,tong2021directed,ding2020phraseclick}, the processing of multi-modal information is drawing more attention from researchers. Referring image segmentation, which involves both vision and language, is one of the most popular topics in this area \cite{wu2020phrasecut,ding2021vision,hu2016natural}. Given an image and a natural language expression referring to any one of the instances in the image, the goal of referring image segmentation is to locate the target instance and generate a mask for it. 

In order to select the target instance from all instances in the image, one of the most challenging problems is to model the interaction among all candidates. For example, given an image of several persons standing and a query expression saying ``the rightmost person'', the model must do comparison among all instances of ``person'' before determining which is the ``rightmost''. Most of the current works use \textit{one-stage methods} \cite{hu2016segmentation,luo2020multi,ye2019cross}, which directly fuse the vision and language features together and perform pixel-level classification on the fused features to generate the segmentation mask. This kind of methods are direct and efficient, but their referring capability is limited as they are instance-agnostic. Without explicitly detected instances, it is hard to directly model and represent the ``interaction'' among them. Thus, this kind of method often does not work as expected in the scenario where the layout of the object is tangled or the input language expression describes a complicated relationship among different instances.

\footnotetext{Fig.~1 and Fig.~2 use ground-truth masks for better illustration. Please find our sample outputs in Section IV.}

To alleviate this issue, \cite{yu2018mattnet} proposes \textit{two-stage methods}, which first detect and generate masks for all instances as proposals with an off-the-shelf instance segmentation method, \eg, Mask R-CNN \cite{he2017mask}, then select the best match among them. They directly model the interaction among instances and can generate high-quality masks with the help of the instance segmentation network. However, in this kind of method, the language information is only used in the matching stage for selecting the existing instance masks, and has no influence on the segmentation stage. Therefore, two-stage methods rely heavily on the off-the-shelf instance segmentation network and their performance is limited by the candidate instances.

In our work, we aim to achieve two goals: first, to be instance-aware so that to model the interaction among instances directly, and second, to be an integrated one-stage method where the instance selection process and mask generating process is simultaneous and collaborative. Inspired by \cite{yolo, wang2019solo}, in our paper, we use grid boxes to represent instances and identify the target by building interaction among them. Meantime, we employ both vision and language features to generate segmentation masks and identify the right one simultaneously.

In our work, as shown in Fig.~\ref{fig:fig1}, we have an Identification Branch that evenly divides the image spatially into a grid of $S\times S$. We let each gird represent the specific instance whose center is located at it. Accordingly, we generate a feature vector for each grid, called an Instance-Specific Feature (ISF). One ISF contains all information about the specific instance, such as size, texture, and shape. ISFs are spatially organized with regard to the spatial location of their corresponding instances into a feature map. This enables the interaction among instances.

To model the interactions among instances, some two-stage methods like \cite{yu2018mattnet} first select a few instances using pre-defined rules, \eg , 5 instances from the same semantic category, and then model the relationship within selected items. Although such artificial selection is reasonable for some cases, it sometimes may not well fit the referring expression. We propose a method that globally models the relationship among all instances by information exchange via instance feature propagation. To this end, a bi-directional propagation process is employed to model the ``comparison'' relationship among different instances. During the propagation, the ISF of each instance is exchanged with all other instances, and finally, the target instance is highlighted.

Meanwhile, we have a Segmentation Branch, which generates a mask for every grid simultaneously. We inject the language features in the mask generation process to enhance the features of the target object. Finally, the mask corresponding to the target grid location is selected as the output mask. The mask generation part can work alone as a one-stage method with limited performance, and our experiment shows that its performance can be greatly enhanced with the awareness of other instances and the interaction modeling among instances. Furthermore, we propose a refinement module that refines the coarse mask to generate a more detailed mask prediction.

In summary, the major contribution or our work can be listed as follows:
\begin{itemize}
   \item We propose a novel referring segmentation framework that generates segmentation masks and identifies the target-of-interest simultaneously and collaboratively by modeling the relationship among all explicitly-defined instances in the image.
   \item We further propose a propagation based Feature Propagation Module (FPM) and a Refinement Module to enhance the performance of the framework for both identifying and segmentation.
   \item Experiments show that the proposed approach outperforms other methods and achieves the new state-of-the-art consistently on all three RefCOCO series datasets.
\end{itemize}

\section{Related Works}

Referring segmentation is one of the most representative tasks of multi-modal information processing, as it combines natural language processing and computer vision. In recent years, a lot of work has been done to solve this problem. These methods can be roughly divided into two categories: One-stage methods like \cite{ye2020dual, chen2019referring} and two-stage methods like \cite{yu2018mattnet}. The basic pipeline of one-stage methods is to fuse the linguistic features and visual features, and directly generate a segmentation mask.
In \cite{hu2016segmentation}, Hu \etal use a recurrent LSTM to extract the linguistic features and a CNN to extract visual features first and then generate a segmentation mask by concatenating and fusing them in a Fully Convolutional Network (FCN)~\cite{long2015fully, ding2019boundary,ding2019semantic}, which shows the feasibility of this kind of methods. Liu \etal~\cite{liu2017recurrent} further adopt a convolutional multimodal LSTM (mLSTM) module to recurrently refine the segmentation result by each word in the query expression. Shi \etal \cite{shi2018key} introduce a query attention model to give each word a weight and a visual context model to map each keyword to different image regions. Ye \etal~\cite{ye2019cross} propose a cross-modal self-attention CMSA module that extracts long-range correspondence between each word in the query expression and the spatial regions of the image. Luo \etal~\cite{luo2020multi} further combine the tasks of referring segmentation and referring comprehension together, building a multi-task network that outputs a bounding box and a segmentation mask together. In contrast, two-stage methods first use instance segmentation methods to detect all instances in the image with their locations and masks, regardless of the query expression, and then tries to give each instance a score measuring how much a certain instance ``matches'' the given query expression. {This kind of method is commonly used in another related task referring expression comprehension \cite{luo2017comprehension, yu2017joint, zhang2017discriminative, wang2019neighbourhood, yang2019dynamic}, which aims to output a bounding box instead of a segmentation mask of the target instance given a referring expression. It is also utilized in the referring expression segmentation, \eg MAttNet \cite{yu2018mattnet}}, which first adopts the Mask R-CNN~\cite{he2017mask} to do instance segmentation and then build a modular network to find the best-match instance.

Unlike the common one-stage methods, our network is enabled to model the interactions of instances directly, which helps the framework perform better on identifying. Different from the two-stage methods that only focus on matching, our unified system interactively and collaboratively learns the vision and language information together, and concerns about the whole referring segmentation process simultaneously.

\begin{figure*}[t!]
   \begin{center}
      \includegraphics[width=\linewidth]{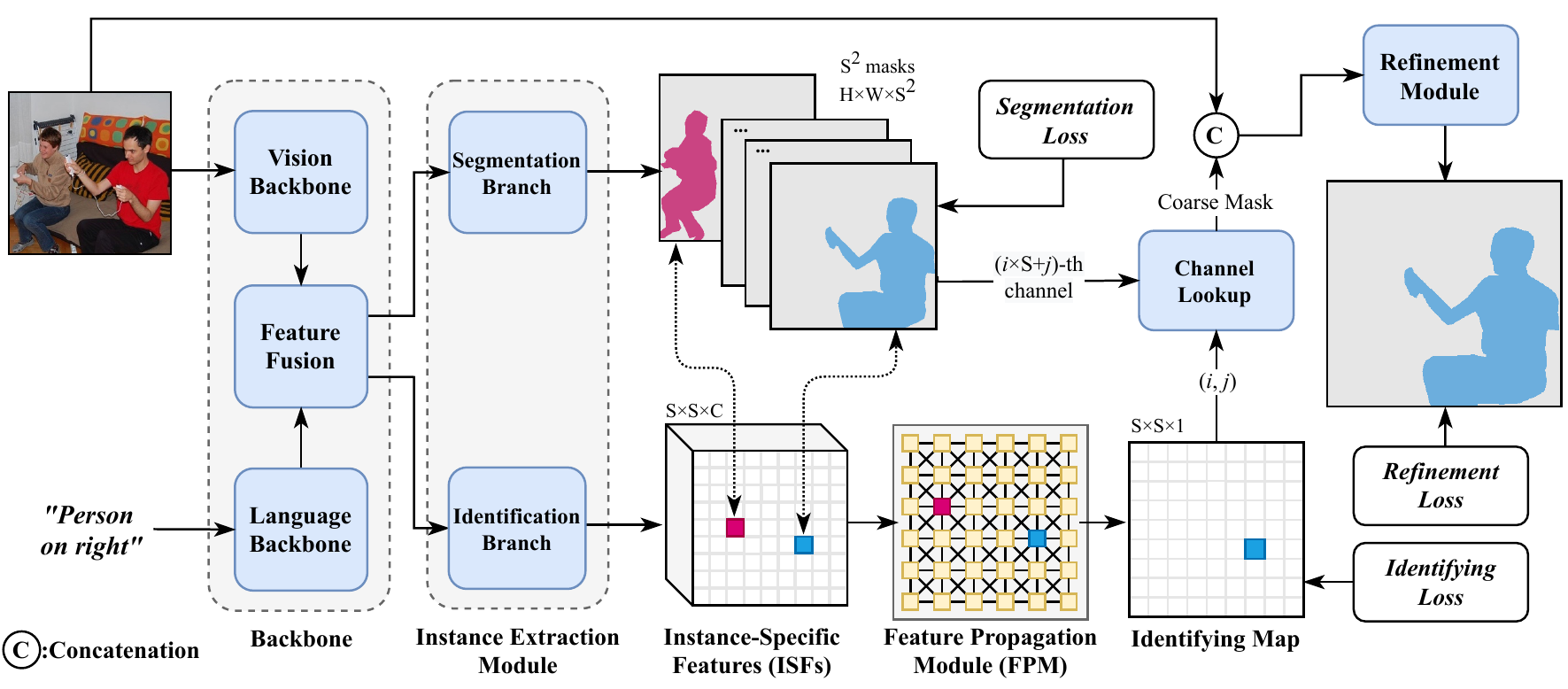}
   \end{center}
   \caption{Overview of the proposed framework. The framework contains four main components: Backbone to extract vision and language features and fuse them together; Instance Extraction Module to extract Instance-Specific Features (ISFs) and generate a mask for each grid; Feature Propagation Module to determine the location of the target in grid; Mask Refinement Module to refine the coarse mask generated by the Segmentation Branch.}
   \label{fig:overview}
\end{figure*}

\section{Methodology}

We propose a novel framework for referring segmentation that explicitly extracts a set of specific feature representations for each instance in the image and models the relationships among all instances by feature propagation. At the same time it generates masks for every instances, and identifies the target's mask according to the established relationships among all instances. An overview of our method is shown in Fig.~\ref{fig:overview}. A training sample consists of an image $I\in \mathbb{R}^{H\times W\times 3}$, a referring expression $T=\{w_i\}_{i=1\dots t}$ with the length of $t$, and a segmentation mask of the target object $M$.

\subsection{Backbone}

First, the Backbone extracts features from $I$ and $T$ and fuses them to a set of fused vision-language feature $F$.

\textbf{Vision feature.}~In order to enhance the multi-scale performance, the Feature Pyramid Network (FPN) \cite{lin2017feature,ding2018context} is employed to extract the basic vision features. {We utilize features from the last three blocks of the FPN and denote them, from higher resolution to lower resolution, as: ${F_{vl}}, {F_{vm}}, {F_{vs}}$ with sizes of $H_l\times W_l\times C_v, H_m\times W_m\times C_v$ and $H_s\times W_s\times C_v$, respectively. $C_v$ is the number of vision feature channels. }

\textbf{Language feature.~}{We first utilize the GloVE \cite{pennington2014glove} to generate embeddings $E$ for all words in the input referring expression.} Then a bi-directional GRU \cite{chung2014empirical} module is applied on $E$ to extract the language features. Suppose the hidden states of all words are $F_w \in \mathbb{R}^{t\times C_l}$, where $C_l$ is the number of language feature channels and $t$ is the number of words in the expression. A set of attention weights { $W_w \in \mathbb{R}^{t\times 1}$ }is derived using a self-attention module to measure the importance of each word in the expression. The final linguistic feature $F_t$ is the weighted sum of all hidden state outputs of the GRU, \ie, $F_t = {W_w}^T F_w$.

\textbf{Feature fusion.~}Next, we generate the fused vision and language features, denoted as $F$. We use a dense multiplication operation for fusion. We first apply a linear layer on language feature $F_t$ and a $1\times 1$ convolutional layer on each of the 3 FPN layers to transform them to the same depth in channel, then do the element-wise multiplication. Let the $f_v^{i,j}\in \mathbb{R}^{1\times C_v}$ denote the feature vector at spatial position $(i,j)$ of ${F_{vl}}, {F_{vm}}$ and ${F_{vs}}$. The fused feature at the pixel $(i,j)$, $f^{i,j}$ is derived as:
\begin{equation}
   {f}^{i,j} = (f_v^{i,j}W_v) * (F_tW_t)
\end{equation}
{where $W_v\in \mathbb{R}^{C_v\times C_f}$ and $W_t\in \mathbb{R}^{C_l\times C_f}$ are learnable parameters, $C_f$ is the number of fused feature channels and $*$ denotes element-wise multiplication. The same fusing process is done on all the three FPN layers ${F_{vl}}, {F_{vm}}$, and ${F_{vs}}$.}

\begin{figure}[t]
   \begin{center}
      \includegraphics[width=\linewidth]{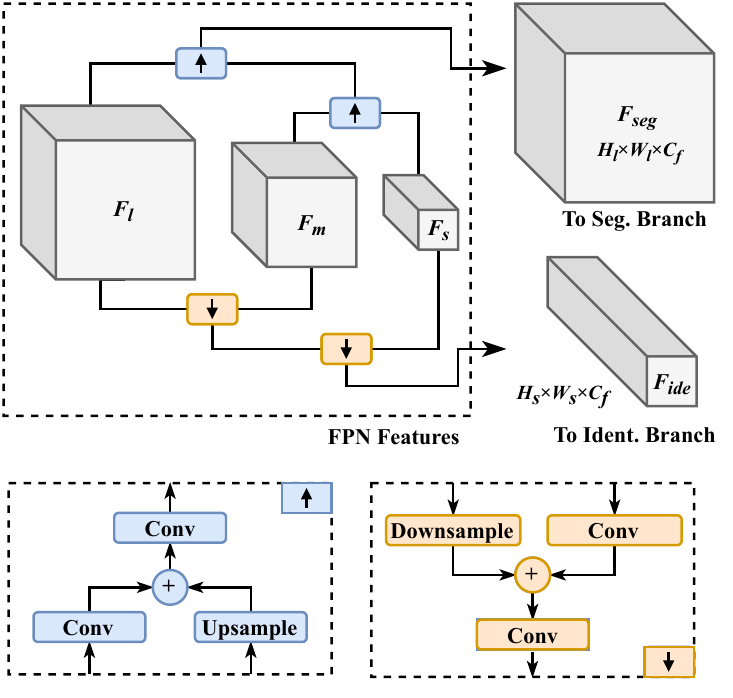}
   \end{center}
   \caption{Three layers of FPN features are fused in two directions. The output of the upsampling pathway is sent to the Segmentation Branch and the output of the downsampling pathway is sent to the Identification Branch.}
   \label{fig:backbone}
\end{figure}

In referring segmentation, it is essential to consider the interaction among different objects, including objects of different sizes. Thus, we further fuse feature from all FPN layers together to enable features exchange across FPN layers, as shown in Fig.~\ref{fig:backbone}. We use two pathway directions based on the lateral connection \cite{lin2017feature} for processing features: a downsampling pathway generating a distilled semantic information output $F_{ide}$ of size $H_s\times W_s$ for the Identification Branch and an upsampling pathway generating a higher resolution feature map $F_{seg}$ of size $H_l\times W_l$ for Segmentation Branch.

\subsection{Instance Extraction Module}

\begin{figure}[t]
   \begin{center}
      \includegraphics[width=\linewidth]{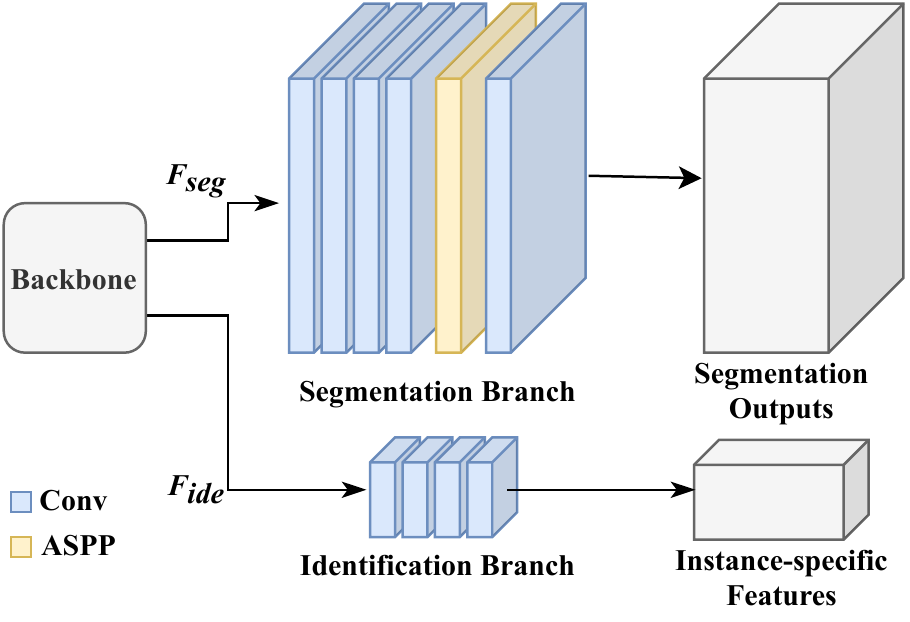}
   \end{center}
   \caption{Architecture of the Instance Extraction Module. The module consists of two branches: an Identification Branch outputting the Instance-Specific Feature (ISF) map where each grid represents an instance and a Segmentation Branch generating masks for each instance.}
   \label{fig:iem}
\end{figure}

The Instance Extraction Module has two branches: Identification Branch and Segmentation Branch. The structure of this module is concise and direct, as shown in Fig.~\ref{fig:iem}.

\textbf{Identification Branch} divides the image into grids and link each instance in the image to its nearest gird. Inspired by \cite{wang2019solo, yolo}, we divide the image into $S\times S$ grid, and each grid is responsible for detecting a possible object centered at its location. Accordingly, the Identification Branch outputs an Instance-Specific Feature (ISF) map $F_{ins}\in \mathbb{R}^{S\times S\times C}$, or $F_{ins}=\{f_{ins}^{i,j}\}_{i,j=0,1,\dots S-1}$, where $C$ is the number of instance feature channels. Since each $f_{ins}^{i,j}$ specifies the characteristics of an object centered at location $(i,j)$, we call each $f_{ins}^{i,j}$ ISF vector.  

The branch takes as input $F_{ide}$ from the backbone, and resize its spatial size to $S\times S$ using interpolation to match the spatial correspondence between feature map and grids. The body is formed by four $3\times 3$ stacked convolutional layers, generating the ISF map $F_{ins}$, where each position corresponds to one grid in the image.

\textbf{Segmentation Branch} takes $F_{seg}$ as input, and generates one binary mask for each grid, leading to a $H\times W\times S^2$ output. To ensure that each ISF vector is uniquely associated with one specific object, we designate a fixed mapping relationship between the spatial location of the grid and the channel location of the target mask: for the grid at the $i$-th row and $j$-th column, the channel number $c$ of its mask will be $i\times S + j$, where $0\le i,j\le S-1, 0\le c\le S^2-1$.

This branch has a concise architecture. Its main body is formed by four $3\times 3$ convolutional layers. Then we use an ASPP \cite{chen2017deeplab} module to enhance its multi-scale performance. At last, we use a convolutional layer with $S^2$ output channels to generate a mask for every grid.

\subsection{Feature Propagation Module (FPM)}

\begin{figure}[t]
   \begin{center}
      \begin{subfigure}[t]{\linewidth}
         \centering
         \includegraphics[width=\linewidth]{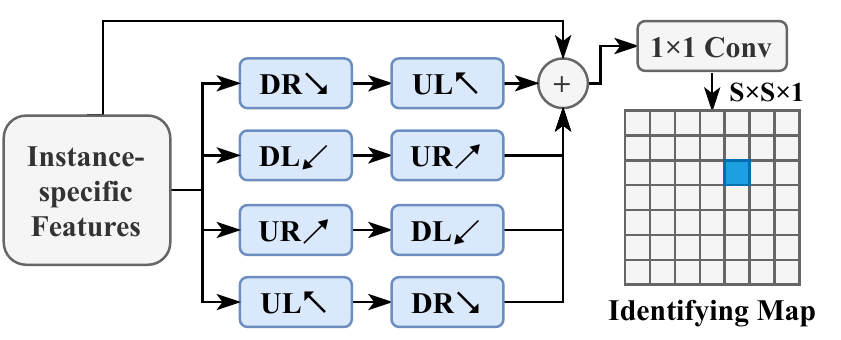}
         \caption{}
         \label{fig:uag-a}
      \end{subfigure}%
      \hspace{0.03\linewidth}%
      \begin{subfigure}[t]{0.45\linewidth}
         \centering
         \includegraphics[width=0.9\linewidth]{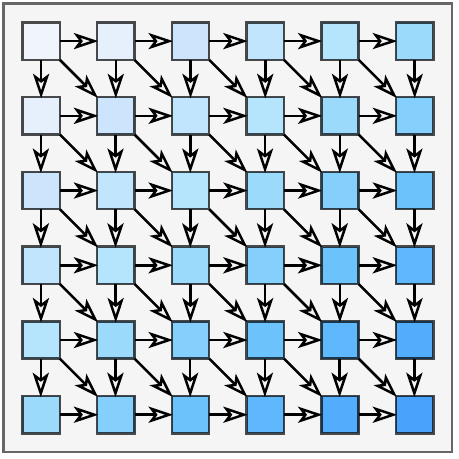}
         \caption{}
         \label{fig:uag-b}
      \end{subfigure}%
      \hspace{0.02\linewidth}%
      \begin{subfigure}[t]{0.45\linewidth}
         \centering
         \includegraphics[width=0.9\linewidth]{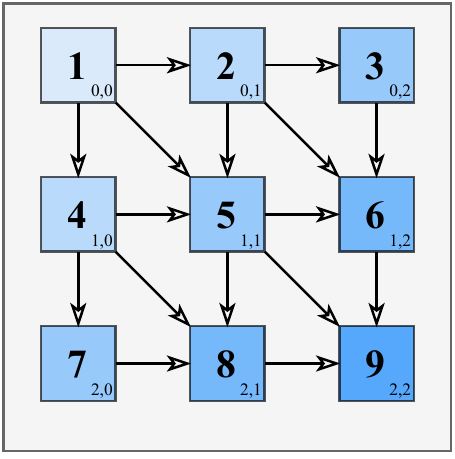}
         \caption{}
         \label{fig:uag-c}
      \end{subfigure}
   \end{center}
   \caption{The Feature Propagation Module. (a): Structure of the four bidirectional paths. The module has four paths, each of which consists of two opposite directions. (b): The detailed propagation process of the DR$\searrow$ (Down, Right) direction. (c):The propagation order in the DR$\searrow$ direction.}
   \label{fig:uag}
\end{figure}

As mentioned before, given an image, a natural way to point out a target is by ``comparison''. To model this relationship, it is essential to let each instance gain knowledge about other instances. From this point, we propose a global modeling method, Feature Propagation Module (FPM), that exchange information among all instances, letting each of them be aware of all others.

The FPM is a recurrent based method, which has four propagation \textit{directions}: DR$\searrow$, DL$\swarrow$, UR$\nearrow$, UL$\nwarrow$ (U, D, L, R: Up, Down, Left, Right) like in Fig.~\ref{fig:uag-b}. For each pixel in one direction, information about the current pixel and the past pixels are fused together, producing a hidden state feature $h$. Then, this feature is sent to the next pixel, as the information about past pixels. The propagation process shares some similarities to RNNs, which starts from one certain pixel and moves towards a specific direction till the last pixel in this direction. Further, in order to ensure that one pixel can gain information from all others, the FPM builds four bidirectional propagation \textit{paths}, each of which consists of two opposite directions, as shown in Fig.~\ref{fig:uag-a}. 

In the beginning of the first propagation direction, the hidden state features are initialized with the Instance-Specific Features (ISF) at their corresponding locations. Since each ISF represents one specific instance in the image, it is convenient to model the interactions among them. For each propagation direction, the hidden state of one pixel updates itself by its past value and at most 3 past pixels: row, column, diagonal, like shown in Fig.~\ref{fig:uag-c}. In each step of propagation, denote the feature vector at position $(i, j)$ as $v^{i,j}$, the hidden state of the position $(i,j)$, $h^{i,j}$, is updated as:
\begin{equation}
   h^{i,j}=
      v^{i,j}+\alpha W_2\sum_{m, n}^{incoming}{h^{m,n}}
\end{equation}
where $W_2\in \mathbb{R}^{C_v\times C_v}$ are learnable weights, $\alpha$ is a constant to control the forget rate. A non-linear activation function is then applied on the derived $h^{i,j}$. Notably, $v^{i,j}=f^{i,j}_{ins}$ in the first propagation direction in Fig.~\ref{fig:uag-a}, while in the second direction, the feature vector $v^{i,j}$ is the hidden state output from the first direction.
For example, in the first direction (DR$\searrow$) of the top path in Fig.~\ref{fig:uag-a} (DR$\searrow$ - UL$\nwarrow$), when $i=j=1$:
\begin{equation}
   \begin{aligned}
      h^{1,1}=W_1f^{1,1}_{ins}+&\alpha W_2\left(h^{0,0}+h^{0,1}+h^{1,0}\right)
   \end{aligned}
\end{equation}
where $W_1\in \mathbb{R}^{C_v\times C_v}$ are also learnable weights.

With the four paths in the bi-directional FPM, we approximate an all-directional propagation and model the ``comparison'' relationship. \Eg, in the DR$\searrow$ - UL$\nwarrow$ path, assume that there are two instances in the image: $A$ located on the top left corner and $B$ on the bottom right. In the first direction (DR$\searrow$), $B$ gets information about $A$. Then in the opposite direction (UL$\nwarrow$), $A$ gets a second-order knowledge that contains information of $B$ itself and $B$'s information of $A$. In such a way, the relationship of ``comparison'' is modeled.

To maintain the instance correspondence property, we sum up the final state from all paths, as well as the input ISFs together. The result is finally processed by a $1\times 1$ convolutional layer, generating a $S\times S\times 1$ Identifying Map $I$. This map indicates the probability of the instance in different grids being the target instance.

\begin{table*}[ht]
   \centering
   \caption{Ablation study results on the validation set of the RefCOCO. (IEM: Instance Extraction Module, FPM: Feature Propagation Module, RM: Refinement Module)}
     \setlength{\tabcolsep}{5mm}{\begin{tabular}{l|c|c|c|c|c|c}
     \hline
     Model  & {Pr@0.5} & {Pr@0.6} & {Pr@0.7} & {Pr@0.8} & {Pr@0.9} & {IoU} \\
     \hline\hline
     Baseline   & 59.88 & 50.64 & 38.70 & 20.68 & 3.36  & 52.40 \\
     Baseline+IEM  & 74.53 & 67.80 & 58.09 & 35.58 & 5.73  & 61.47 \\
     Baseline+IEM+FPM & 76.23 & 70.87 & 61.03 & 36.99 & 6.33  & 62.48 \\
     {Baseline+IEM+RM}  & {75.44} & {68.45} & {60.09} & {36.98} & {9.73}  & {63.47} \\
     \hline\hline
     Baseline+IEM+FPM+RM & \textbf{77.18} & \textbf{73.43} & \textbf{66.59} & \textbf{53.01} & \textbf{18.85} & \textbf{65.19} \\
     \hline
     \end{tabular}}%
   \label{tab:ablation}%
 \end{table*}%

\subsection{Refinement Module}

{Once the target probability map is derived, the segmentation mask that corresponds to the grid position $(i, j)$ of the maximum value, \textit{i.e.}, the $(i\times S + j)$-th channel of the Segmentation Branch output, is selected as the target mask. This is indicated as ``Channel Lookup'' block in \figurename{~\ref{fig:overview}}}. However, because the output spatial size of the Segmentation Branch is limited due to the limitation of computational resources, and the vision features used in our network are from higher-level layers of the vision backbone, the output directly taken from the Segmentation Branch is coarse. It is desired to introduce low-level and high-solution features to enhance the spatial details of predicted mask. Thus, we concatenate the original image and the resized predicted mask together, as the input to the refinement module. The refinement module consists of three $3\times3$ convolution layers with upsampling layers in between and outputs a 1-channel prediction map. This output map is then added with the upsampled coarse segmentation map from the Segmentation Branch to form the final prediction.

\subsection{Loss Functions}

In the proposed network, there are three loss terms to guide the network training: identifying loss $l_{ide}$, segmentation loss $l_{seg}$, and refinement loss $l_{ref}$. The places where these losses are applied are shown in Fig.~\ref{fig:overview}. The whole loss function is formed by the weighted sum of all loss terms:

\begin{equation}
   l=w_{ide}l_{ide}+w_{seg}l_{seg}+w_{ref}l_{ref}
\end{equation}

\textbf{Identifying loss.} The ground-truth of identifying loss is a $S\times S \times 1$ grid map, indicating the grid location of the target instance. We first calculate the gravity center $G$ of the target instance, then label the grid that $G$ falls into and its 8 neighborhood grids by 1, called positive grids $G_{pos}$. All other grids are annotated as negative grids with 0. The loss is applied on the Identifying Map $I$, \ie the output of the FPM.
\begin{equation}
    y_{ins}^{i,j}=\left\{
      \begin{aligned}
         &1, &(i, j)\in G_{pos}\\
         &0, &Otherwise
      \end{aligned}
    \right.
\end{equation}
\begin{equation}
l_{ins} = \text{BCE}(y_{ins}, I)
\end{equation}
The Binary Cross Entropy (BCE) is used to measure the difference between the predictions and ground-truths. We use BCE in all three loss terms.

\textbf{Segmentation Loss.} The Segmentation Branch outputs $S^2$ masks, each of which corresponds to one grid of the Identification Branch. The ground-truth corresponding to the positive grid is the ground-truth instance mask. Notably, in referring segmentation, we only have one mask for each training sample: the mask of the instance that is referred to by the referring expression. Thus, we set the loss for negative grids to be zero and do not use them in training.
\begin{equation}
    l_{seg}=\left\{
      \begin{aligned}
         &\text{BCE}(y_{seg}^{iS + j}, M), &(i, j)\in G_{pos}\\
         &0, &Otherwise
      \end{aligned}
    \right.
\end{equation}

\textbf{Refinement Loss.} The refinement loss is the BCE of the ground-truth mask and the predicted mask. In the early stage of training, the output of the Identification Branch will be very inaccurate. Since the gradient of the refinement branch can be propagated back, selecting the wrong channel in the Segmentation Branch will be harmful to the whole network. Thus, we use an adaptive training strategy in this module. During training, after the coarse segmentation map is selected from the Segmentation Branch output, we evaluate its IoU with the ground-truth. If the coarse IoU is greater than a threshold $\theta$, it is considered to be a successful {identification}. Otherwise, we consider the channel selection is wrong and set the refinement loss as 0.
\begin{equation}
    l_{ref}=\left\{
      \begin{aligned}
         &\text{BCE}(x_{ref}, M), &IoU_{coarse}\ge \theta\\
         &0, &IoU_{coarse}< \theta
      \end{aligned}
    \right.
\end{equation}

\begin{table}[t!]
   \caption{Performance gap of replacing the output of each branch with the Ground-Truth. The experiments reveal the great potential of our new framework.}
   \setlength{\tabcolsep}{2.2mm}{\begin{subtable}[]{\linewidth}
      \centering
      \caption{Replacing the output of Identification Branch with GT}
      \begin{tabular}{l|c|c|c}
         \hline
               & {GT Location}    & {Predicted Location}  & Gap      \\
         \hline\hline
         val   & $76.32$ & $65.19$ & $-10.13$ \\
         testA & $78.23$ & $68.45$ & $-9.78$ \\
         testB & $73.35$ & $62.73$ & $-10.62$ \\
         \hline
      \end{tabular}%
      \label{tab:gtid}%
   \end{subtable}}
   \vspace{0.15in}
   
   \setlength{\tabcolsep}{3.4mm}{\begin{subtable}[]{\linewidth}
      \centering
      \caption{Replacing the output of Segmentation Branch with GT}
      \begin{tabular}{l|c|c|c}
         \hline
               & {GT Mask}    & {Predicted Mask}  & Gap      \\
         \hline\hline
         val   & $80.41$ & $65.19$ & $-14.22$ \\
         testA & $82.48$ & $68.45$ & $-14.03$ \\
         testB & $77.22$ & $62.73$ & $-14.49$ \\
         \hline
      \end{tabular}%
      \label{tab:gtmask}%
   \end{subtable}}%
   \label{tab:gtbranch}
\end{table}%

\begin{table*}[t!]
   \centering
   \caption{Experimental results of the IoU metric, and comparison of other methods with ours. $^*$: Google split. Type: I: One-stage methods; II: Two-stage methods.}
   {\begin{tabular}{l|c|c|c|c|c|c|c|c|c|c|c}
      \hline
      \multirow{2}[4]{*}{Model} & \multirow{2}[4]{*}{Type} & \multirow{2}[4]{*}{Backbone} & \multicolumn{3}{c|}{RefCOCO} & \multicolumn{3}{c|}{RefCOCO+} & \multicolumn{3}{c}{RefCOCOg} \\
      \cline{4-12}
                                     & & & val   & testA & testB & val   & testA & testB & val   & test  & val$^*$\\
      \hline\hline
      DMN \cite{margffoy2018dynamic} & I  & DPN92 & 49.78 & 54.83 & 45.13 & 38.88 & 44.22 & 32.29 & -     & -     & 36.76 \\
      RRN \cite{li2018referring}     & I  & DeepLab101 & 55.33 & 57.26 & 53.93 & 39.75 & 42.15 & 36.11 & -     & -     & 36.45 \\
      CMSA \cite{ye2019cross}        & I  & DeepLab101 & 58.32 & 60.61 & 55.09 & 43.76 & 47.60 & 37.89 & -     & -     & 39.98 \\
      CMPC \cite{huang2020referring} & I  & DeepLab101 & 61.36 & 64.53 & 59.64 & 49.56 & 53.44 & 43.23 & -     & -     & 49.05 \\
      LSCM \cite{hui2020linguistic}  & I  & DeepLab101 & 61.47 & 64.99 & 59.55 & 49.34 & 53.12 & 43.50 & -     & -     & 48.05 \\
      MCN \cite{luo2020multi}        & I  & Darknet53 & 62.44 & 64.20 & 59.71 & 50.62 & 54.99 & 44.69 & 49.22 & 49.40 & -     \\
      CGAN \cite{luo2020cascade}     & I  & Darknet53 & 64.86 & 68.04 & 62.07 & 51.03 & 55.51 & 44.06 & 51.01 & 51.69 & 46.54 \\
      \hline
      MAttNet \cite{yu2018mattnet}   & II & MaskRCNN-R101 & 56.51 & 62.37 & 51.70 & 46.67 & 52.39 & 40.08 & 47.64 & 48.61 & -     \\
      NMTree \cite{ye2019cross}      & II  & MaskRCNN-R101 & 56.59 & 63.02 & 52.06 & 47.40 & 53.01 & 41.56 & 46.59 & 47.88 & -    \\
      CAC \cite{chen2019referring}   & II & MaskRCNN-R101 & 58.90 & 61.77 & 53.81 & -     & -     & -     & -     & -     & 44.32 \\
      \hline\hline
      Ours & - & Darknet53 & \textbf{65.19} & \textbf{68.45} & \textbf{62.73} &\textbf{52.70} & \textbf{56.77} & \textbf{46.39} & \textbf{52.67} & \textbf{53.00} & \textbf{50.08} \\
      \hline
   \end{tabular}}%
   \label{tab:iou}%
\end{table*}%

\begin{table*}[t]
    \centering
    \caption{Experimental results of the Precision metric, and comparison of other methods with ours on the validation set of the RefCOCO.}
    \setlength{\tabcolsep}{3.6mm}{\begin{tabular}{l|c|c|c|c|c|c}
     \hline
     Model & {Pr@0.5} & {Pr@0.6} & {Pr@0.7} & {Pr@0.8} & {Pr@0.9} & {IoU} \\
     \hline\hline
     STEP \cite{chen2019see}  & 70.15 & 63.37 & 53.15 & 36.53 & 10.45 & 59.13 \\
     CMSA \cite{ye2019cross}  & 66.44 & 59.70 & 50.77 & 35.52 & 10.96 & 58.32 \\
     BRINet \cite{hu2020bi} & 71.83 & 65.05 & 55.64 & 39.36 & 11.21 & 61.35 \\
     LSCM \cite{hui2020linguistic} & 70.84 & 63.82 & 53.67 & 38.69 & 12.06 & 61.54\\
     CMPC \cite{huang2020referring}  & 71.27 & 64.44 & 55.03 & 39.28 & \textbf{12.89} & 61.19 \\
     MCN \cite{luo2020multi}  & \textbf{76.60} & 70.33 & 58.39 & 33.68 & 5.26  & 62.44 \\
     \hline\hline
     Ours  & \textbf{76.60} & \textbf{72.15} & \textbf{65.06} & \textbf{47.66} & 11.85 & \textbf{65.67} \\
     \hline
     \end{tabular}}%
    \label{tab:prec}%
 \end{table*}%

\section{Experiments}

\subsection{Implementation Details}

We train and test the proposed approach on three widely-used referring segmentation datasets: RefCOCO~\cite{yu2016modeling}, RefCOCO+~\cite{yu2016modeling} and RefCOCOg~\cite{mao2016generation, nagaraja2016modeling}. Since the Instance Extraction Module needs to find different instances in the image and product instance-level segmentation mask for multiple instances, instance-aware ability is desired for our model. Therefore, we chose the Darknet backbone following the experiment settings in the previous works~\cite{yu2018mattnet,luo2020multi}. All test/val images in three RefCOCO datasets are excluded when training the Darknet backbone. We follow prior work \cite{luo2020multi,huang2020referring} and adopt the GloVe word embeddings \cite{pennington2014glove} on Common Crawl 840B tokens. We use Adam with the base learning rate of $0.001$ for optimization. Images are resized to $416\times 416$ before sending to the network, and the network outputs to 8x downsampled mask to save memory usage in our implementation. All convolutional layers, except for the output layer of each module, are followed with a Leaky ReLU activation function and batch normalization. We set $w_{ide}=10.0,~w_{seg}=0.03$,~$w_{ref}=0.5$,~$\theta=0.3$, and grid number $S\times S=13\times 13=169$ in our settings. The network is trained and evaluated on four 11G Nvidia RTX 2080Ti GPUs and two Intel Xeon 4112 CPUs with the batch size of $12$, and the training process takes 16 hours for 55 epochs.

\subsection{Ablation Study}

In this section, we report the ablation study results of our framework. Besides the regular IoU, we also use another widely-used metric: Precision@X, to better demonstrate the function of each module in our framework. The Precision@X represents the percentage of images whose prediction IoU is above threshold X. The value of low-threshold precisions reflects the percentage of successfully identified instances, which corresponding to the identifying ability of the model. The value of high-threshold precisions reflects the quality of output masks, corresponding to the model's segmentation ability. Following \cite{hu2020bi,huang2020referring,luo2020multi}, we compare the precision from threshold 0.5 to 0.9, on the validation set of the RefCOCO dataset. The results are shown in Table~\ref{tab:ablation}.

For our baseline model, we remove the Identification Branch, and change the output channel number of the Segmentation Branch to 1, \ie, let the Segmentation Branch directly output the mask of the target instance, which is the same as one-stage methods. From Table~\ref{tab:ablation}, we can see that our baseline, with the Segmentation Branch only, works well as a one-stage method and achieves acceptable results. However, its precision at low-threshold is not high, which indicates its defects in identifying. 

For the ``Baseline+IEM'' model, we recover the Identification Branch and Segmentation Branch in the Instance Extraction Module (IEM), making network instance-aware, but replace the Feature Propagation Module (FPM) with a convolutional layer. It can be seen that the performance is significantly increased, especially at lower thresholds, which shows that the IEM greatly enhances the identifying ability of the network. 

Next, we test the proposed FPM. The proposed FPM further improves the IoU performance by enhancing the identifying ability. The improvements brought by FPM shows that it can further enhance prediction by reasoning relationships among instances in the input image. {Also, from the last two rows of Table~\ref{tab:ablation}, we can see that the Refinement Module (RM) enhances the network's performance more significantly at higher thresholds. Comparing the model ``Baseline+IEM'' with RM and without RM, a performance gain of about $4\%$ can be get for the Pr@0.8 and Pr@0.9 metrics.} This demonstrates the effectiveness of the RM on enhancing segmentation mask details. 

Compared to the baseline, the overall performance of the proposed framework is 13.79\% better in terms of IoU and 17.30\% better in terms of Pr@0.5. This demonstrates the superiority of our method and shows that our method significantly enhances the referring segmentation from different aspects, \ie, identifying, and segmentation.

\begin{figure*}[ht]
   \begin{center}
      \includegraphics[width=\linewidth]{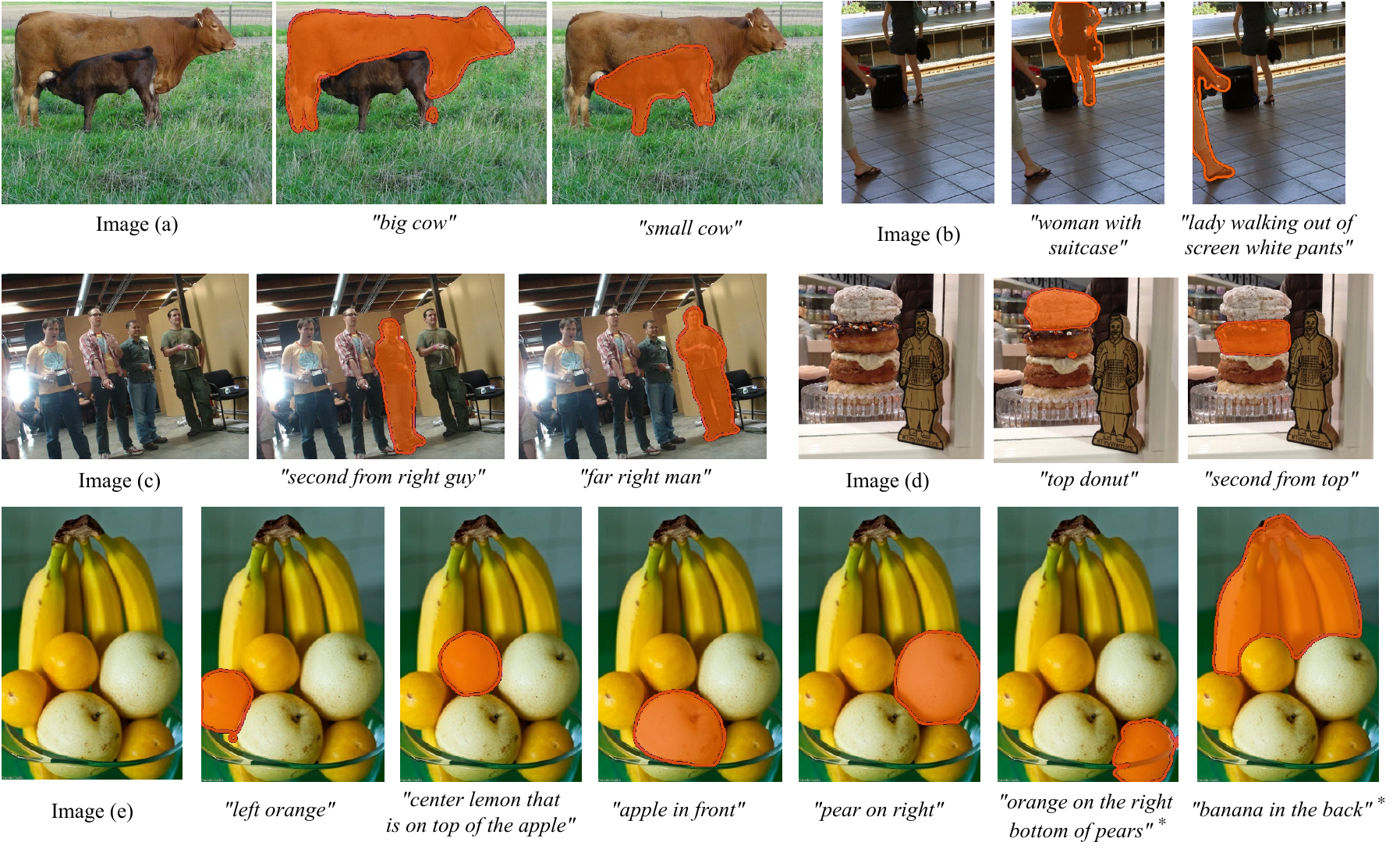}
      \caption{Qualitative examples of the proposed approach. The segmented part is highlighted in orange. (*: The datasets do not provide referring expressions for these two instances. We create phrases for them as user study.)}
      \vspace{-0.05in}
      \label{fig:demo}
   \end{center}
\end{figure*}

\input{images/qual_compare}

\input{images/fpm_output}

\input{images/mask_output}

\begin{figure*}
   \begin{center}
         \begin{subfigure}[t]{0.44\linewidth}
         \centering
         \includegraphics[height=0.32\linewidth]{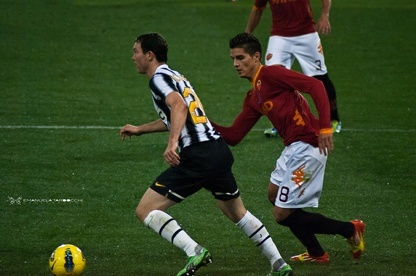}
         \includegraphics[height=0.32\linewidth]{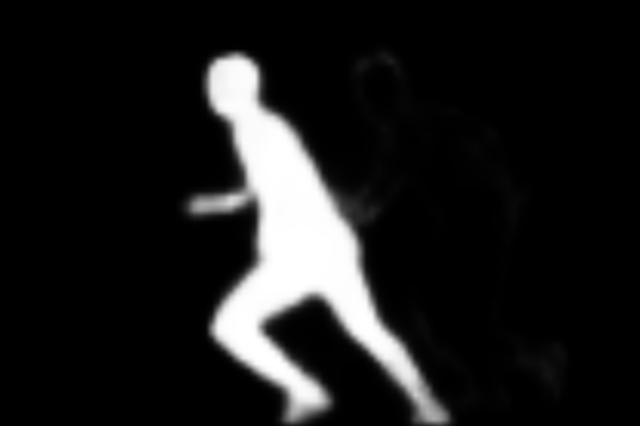}
         \caption{\textit{``8''}}
         \label{fig:fail1}
      \end{subfigure}%
      \hspace{0.02\linewidth}
      \begin{subfigure}[t]{0.44\linewidth}
         \centering
         \includegraphics[height=0.32\linewidth]{images/failure_case/fail30.png}
         \includegraphics[height=0.32\linewidth]{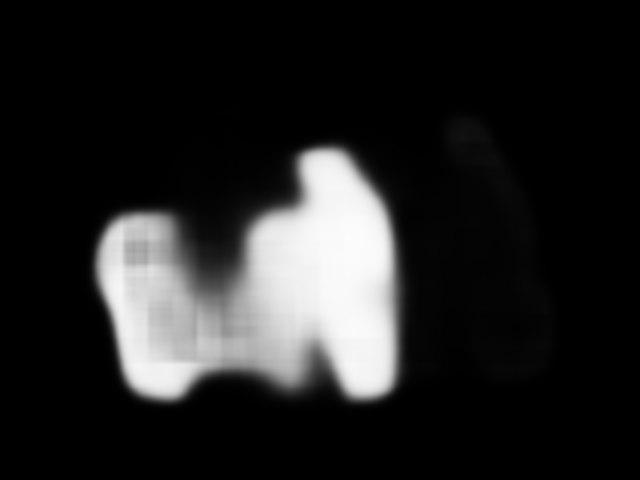}
         \caption{``closest motorcycle not rider''}
         \label{fig:fail2}
      \end{subfigure}
   \end{center}
   \vspace{-0.1in}
  \caption{Visualization of some failure cases and input language expressions. For each sample, left: input image; right: predicted mask.}
   \label{fig:fail}
\end{figure*}

\subsection{Branch Performance}

In this section, we analyze the detailed performance of each branch in Table~\ref{tab:gtbranch}, by using the ground-truth to take over its output. The gap between the resulting performance and the original performance shows the areas for continued development of this branch. Notably, FPM is included in the Identification Branch in experiments in this section.

\textbf{Identification Branch:~Prediction~\vs~GT.} We disable the Identification Branch and use the ground-truth identify map to substitute for its output. This makes the model always find the right grid location of the target instance, which represents the upper limit of the identifying ability of our model. Results are reported in Table~\ref{tab:gtbranch}~(a).

\textbf{Segmentation Branch:~Prediction~\vs~GT.} In the experiment, we replace all the output masks of Segmentation Branch with the corresponding ground-truth masks. In other word, the network always outputs the ground-truth mask of the object selected by the Identification Branch, whether this selection is correct or not. We get higher results with the ground-truth mask (see Table~\ref{tab:gtbranch}~(b)), which demonstrate the effectiveness of the Identification Branch.

Comparing the two experiments, it is shown that replacing the outputs of Segmentation Branch to the ground-truth gives more performance gain. This is because the scale of the Segmentation Branch in our approach is small and the architecture is direct.

\subsection{Results on Benchmarks}

Here we compare the proposed approach with previous state-of-the-art methods on three public referring segmentation benchmarks, RefCOCO, RefCOCO+, and RefCOCOg. We report the IoU results of the proposed approach against other methods in Table.~\ref{tab:iou}. It can be seen that our approach outperforms previous state-of-the-art methods on three datasets by $\sim 1\%$ in term of IoU. On the testB set of RefCOCO+, the performance of our method is over $2\%$ higher than the previous state-of-the-art method, CGAN\cite{luo2020cascade}. On the test set of the RefCOCOg - UMD split, the performance of our method is also about $1.5\%$ superior to the previous works. 

In Table~\ref{tab:prec}, we compare the Precision metric of our method other methods that have data available. It can be seen that in terms of the targeting performance, our method performs the same as MCN. Considering that MCN uses more data to train (mask + bounding box) while our method only uses mask to train, it is reasonable to say that our method better identifies the target than other previous methods. Besides, our method is achieves the highest score on Pr@0.6, 0.7 and 0.8.

\subsection{Visualizations}

\textbf{Example Results} of our method on the val/test split of three datasets are shown in Fig.~\ref{fig:demo}. To demonstrate the ability of our method of analyzing the relationship between instances, we report the segmentation results of different referring expressions on each image. Image (a) of Fig.~\ref{fig:demo} shows that our method is aware of the spatial sizes of instances, and in image (b) our method models instances using attributes information. In image (c) and (d), we show more complicated cases where exist numbered relationships, \eg, ``second from right''. From the figure, we can see that our method successfully targets the instance with an expression that describes only the relative locations without the help of other information such as attributes. This shows that our method has a strong ability on modeling the relative relationship among objects. The image (e) is the most difficult case where multiple instances of different semantic categories are tangled together in a complex layout. Notably, the dataset only provides referring expressions for at most four instances in the image, so we add two more expressions of the rest of the instances in the last of the figure for a complete demonstration. Our method makes accurate identification of all instances in the image. It can be seen that our method can handle the relationship between objects of different semantic categories, \eg \textit{center lemon that is on top of the apple}, and can identify objects of different sizes, \eg the big banana in the back, as well as small oranges in the front.

\figurename{~\ref{fig:qual_comp}} shows the qualitative comparison among our method, the baseline model as defined in the ablation study section, and the previous state-of-the-art model, MCN \cite{luo2020multi}. The first case describes a relationship between two objects. The baseline and MCN failed to extract the target, but our model successfully segmented the correct one. The second example describes a relationship between two instances of a same semantic category (sandwich). The baseline model failed to find the correct instance in this case. MCN managed to find the target instance, but the quality of the segmentation mask is not good, as a significant part of a non-target instance is included in the output. Our proposed model finds the correct object, and also generates a high-quality segmentation mask. These examples show that our method has a superior performance in analyzing the relationship between instances.

\textbf{Identification Branch.} Fig.~\ref{fig:ramout} shows more example outputs of our model handling referring expressions describing different scenarios such as instance relationships (Image (a)), positions (Image (b)) and attributes (Image (a)), together with the Identifying Map, \ie, the output of the FPM module. It is shown that the FPM does not output segmentation masks, but only highlights the grids that corresponding to the target instance.

\textbf{Segmentation Branch.} As we mentioned previously, we inject the language features to the Segmentation Branch to enhance the features of the target object. Thus, to illustrate the effectiveness of the Segmentation Branch and demonstrate the influence of language features on it, we visualize the output of the Segmentation Branch in Fig.~\ref{fig:eg_mask}. From the figure, it can be seen that the output of the Segmentation Branch is influenced by language features. However, the model is still aware of other non-target objects, only the segmentation quality is influenced in different input expressions. Since the Segmentation Branch has a small network scale that only consists of a few convolutional layers but has a hard task to generate $S^2$ masks, this setting can help the network focus on outputting fine-grained masks for instances that are more possible to be the target. This also shows that though we only use one instance mask for each sample in training, our network still aware of other non-target instances. We also test our model with the languistic feature removed in the Segmentation Branch. In this case, the IoU performance will decrease by $0.48\%$ on the validation set of the RefCOCO dataset.

\subsection{Failure Cases}

In this section, we analyze some examples of failure cases produced by our method. The failure cases can roughly be divided into two categories: wrong identification and inaccuracy. For wrong identification cases, the model does not output the target referred by the language expression but finds something else, therefore the IoU for these cases are very low, usually $<40\%$. In many cases, the reason is that the given language expressions are ambiguous, or have insufficient information. We show an examples in Fig.~\ref{fig:fail1}. In the figure, there is only a number in the language expression, which is over-simplified and does not provide much information. Besides, since a number is given to refer to the target, it requires OCR to identify the target. This is over-complicated for our model. For inaccuracy cases, the model can successfully identify and extract the target from the image, but the quality of the mask is limited. The IoU for these cases are relatively higher at $40\% - 60\%$. This is mainly due to the complexities in the input image. For example in Fig.~\ref{fig:fail2}, two motorcycles with similar and deceptive colors are overlapped and tangled together, which makes it difficult for our network to accurately distinguish each one from the other.

\section{Conclusion}

In this work, we address the challenging task of referring segmentation. To inherit the merits and overcome the limitations of the previous one-stage and two-stage methods, we propose a new framework that simultaneously and collaboratively segments instances in the image and builds the interactions among them for identification. We propose a feature propagation module to model the comparison relationships among all instances. Besides, we propose a refinement module that introduces low-level and high-resolution features to enhance the spatial details of predicted segmentation mask. Experiments show that our approach enhances both the targeting performance and the mask quality. Without bells and whistles, we achieve new state-of-the-art results on three referring segmentation datasets, which demonstrates the effectiveness of our proposed approach.

\ifCLASSOPTIONcaptionsoff
  \newpage
\fi

{\small
\bibliographystyle{IEEEtran}
\bibliography{egbib}
}

\vfill

\end{document}

%% file: images/qual_compare.tex
\begin{figure*}[t]
   \begin{center}
      \captionsetup[subfigure]{labelformat=empty}

      {\begin{subfigure}[t]{0.49\linewidth}
            \stackon{\includegraphics[height=0.86in]{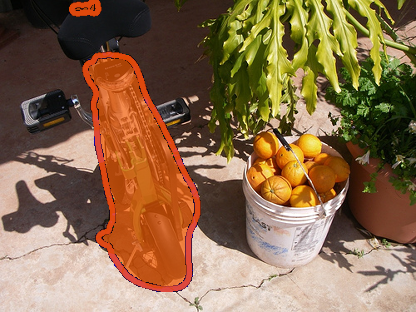}}{Baseline\vphan}%
            \hspace{0.02in}%
            \stackon{\includegraphics[height=0.86in]{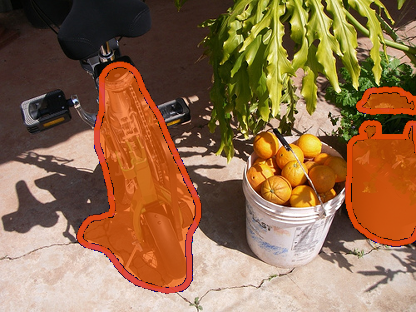}}{MCN~\cite{luo2020multi}\vphan}%
            \hspace{0.02in}%
            \stackon{\includegraphics[height=0.86in]{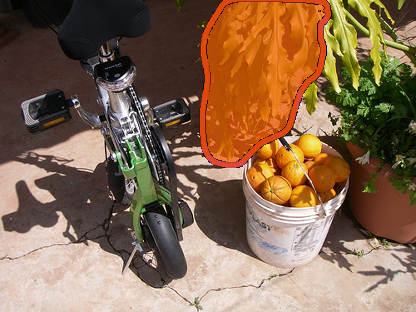}}{Ours\vphan}

            \caption{\textit{Input: ``green hanging over bucket''}}
         \end{subfigure}}%
      \hspace{0.06in}%
      {\begin{subfigure}[t]{0.49\linewidth}
         \stackon{\includegraphics[height=0.86in]{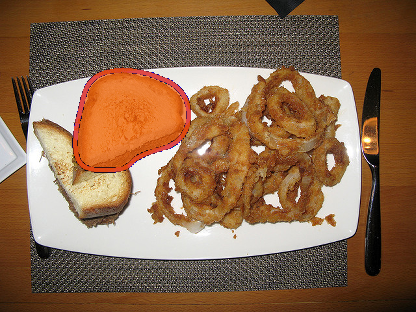}}{Baseline\vphan}%
         \hspace{0.02in}%
         \stackon{\includegraphics[height=0.86in]{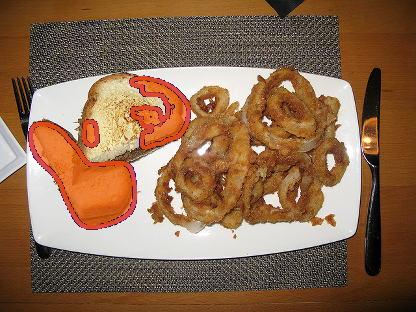}}{MCN~\cite{luo2020multi}\vphan}%
         \hspace{0.02in}%
         \stackon{\includegraphics[height=0.86in]{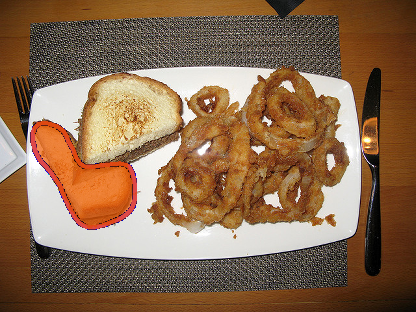}}{Ours\vphan}

         \caption{\textit{Input: ``sandwich half under other one''}}
      \end{subfigure}}
      \caption{Qualitative comparison among the baseline model, previous state-of-the-art method MCN~\cite{luo2020multi}, and our proposed method.}
      \label{fig:qual_comp}
   \end{center}
\end{figure*}

%% file: images/fpm_output.tex
\begin{figure*}[t]
   \begin{center}
      \captionsetup[subfigure]{labelformat=empty}
      \setlength{\fboxsep}{1.5pt}
      \fbox{\begin{subfigure}[t]{1.32in}
            \stackon{\includegraphics[height=0.98in]{images/fpm_out/1_img}}{\vphan}
            \caption{\textit{Image (a)}}
         \end{subfigure}}%
      \hspace{1pt}
      \fbox{\begin{subfigure}[t]{2.64in}
            \stackon{\includegraphics[height=0.98in]{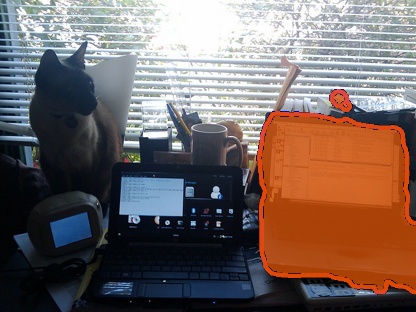}}{Prediction\vphan}%
            \hspace{0.02in}%
            \stackon{\includegraphics[height=0.98in]{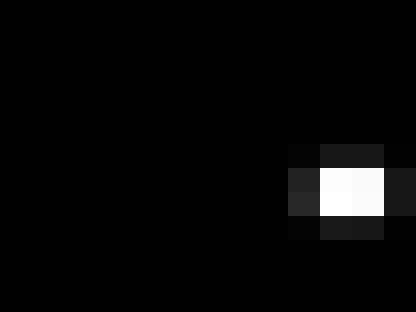}}{Identifying Map\vphan}
            \caption{\textit{Input: ``computer on right side of picture''}}
         \end{subfigure}}%
      \hspace{1pt}
      \fbox{\begin{subfigure}[t]{2.64in}
            \stackon{\includegraphics[height=0.98in]{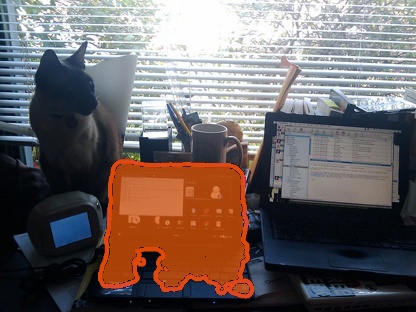}}{Prediction\vphan}%
            \hspace{0.02in}%
            \stackon{\includegraphics[height=0.98in]{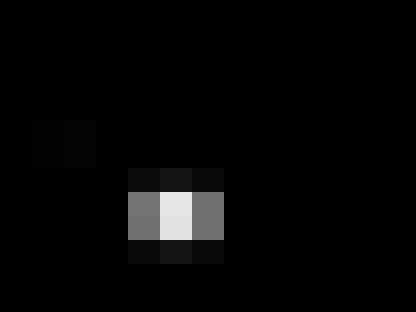}}{Identifying Map\vphan}
            \caption{\textit{Input: ``laptop next to cat''}}
         \end{subfigure}}

      \vspace{0.1in}

      \fbox{\begin{subfigure}[t]{1.32in}
            \stackon{\includegraphics[width=1.309in]{images/fpm_out/2_img}}{\vphan}
            \caption{\textit{Image (b)}}
         \end{subfigure}}%
      \hspace{1pt}
      \fbox{\begin{subfigure}[t]{2.64in}
            \stackon{\includegraphics[width=1.309in]{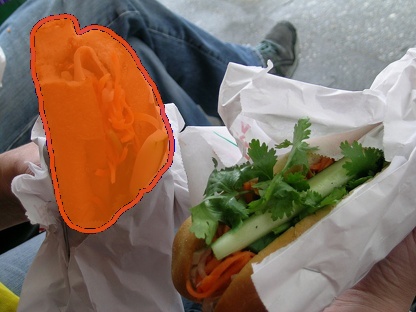}}{Prediction\vphan}%
            \hspace{0.02in}%
            \stackon{\includegraphics[width=1.309in]{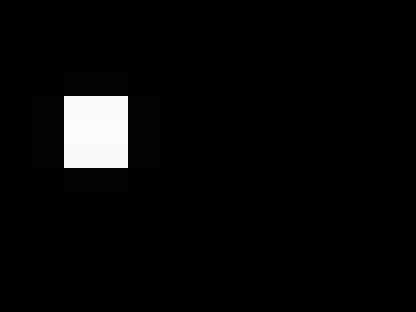}}{Identifying Map\vphan}
            \caption{\textit{Input: ``left sandwich''}}
         \end{subfigure}}%
      \hspace{1pt}
      \fbox{\begin{subfigure}[t]{2.64in}
            \stackon{\includegraphics[width=1.309in]{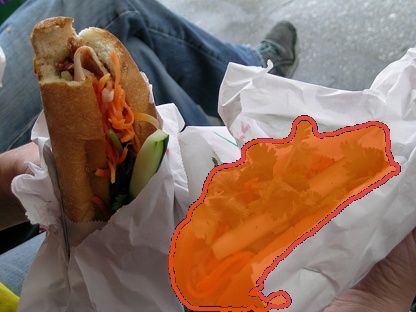}}{Prediction\vphan}%
            \hspace{0.02in}%
            \stackon{\includegraphics[width=1.309in]{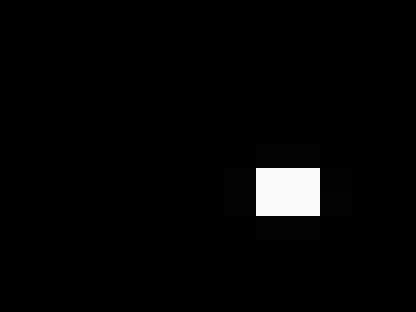}}{Identifying Map\vphan}
            \caption{\textit{Input: ``sandwich on right''}}
         \end{subfigure}}

      \vspace{0.1in}

      \fbox{\begin{subfigure}[t]{1.32in}
            \stackon{\includegraphics[width=1.309in]{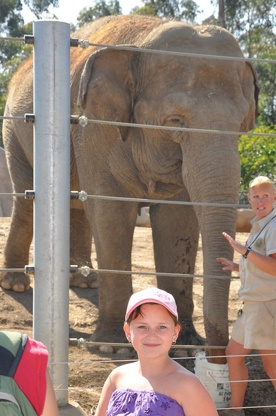}}{\vphan}
            \caption{\textit{Image (c)}}
         \end{subfigure}}%
      \hspace{1pt}
      \fbox{\begin{subfigure}[t]{2.64in}
            \stackon{\includegraphics[width=1.309in]{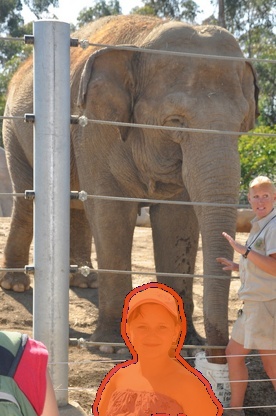}}{Prediction\vphan}%
            \hspace{0.02in}%
            \stackon{\includegraphics[width=1.309in]{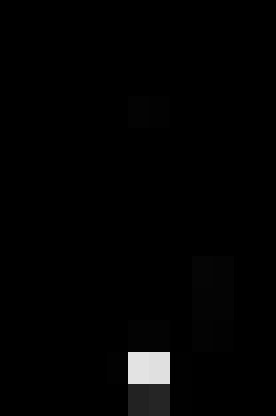}}{Identifying Map\vphan}
            \caption{\textit{Input: ``girl in purple''}}
         \end{subfigure}}%
      \hspace{1pt}
      \fbox{\begin{subfigure}[t]{2.64in}
            \stackon{\includegraphics[width=1.309in]{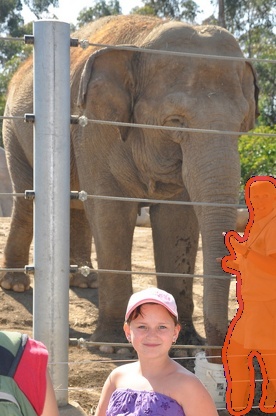}}{Prediction\vphan}%
            \hspace{0.02in}%
            \stackon{\includegraphics[width=1.309in]{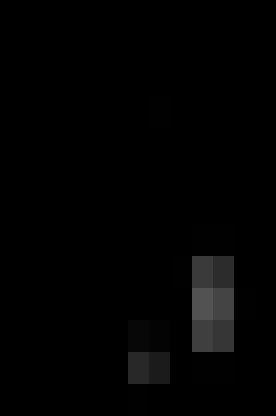}}{Identifying Map\vphan}
            \caption{\textit{Input: ``woman in shorts''}}
         \end{subfigure}}
      \vspace{0.1in}
      \caption{Visualization of the Identification Branch. The ``Identifying Map'' shows the output of the FPM.}
      \label{fig:ramout}
   \end{center}
\end{figure*}

%% file: images/mask_output.tex
\begin{figure*}[t]
    \begin{center}
       \captionsetup[subfigure]{labelformat=empty}
       \setlength{\fboxsep}{1.5pt}
       \begin{center}
          \fbox{
             \begin{subfigure}[t]{1.25in}%
                \stackon{\includegraphics[width=1.23in]{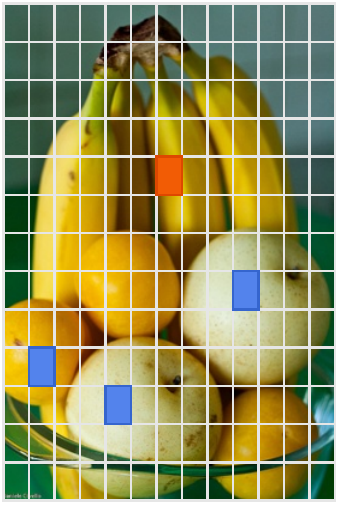}}{ \vphan}
                \caption{\textit{Image}}
             \end{subfigure}}%
          \hspace{0.1in}
          \fbox{
             \begin{subfigure}[t]{5.2in}
                \stackon{\includegraphics[width=1.23in]{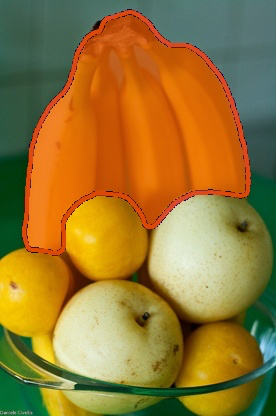}}{\textcolor{orange}{Grid 1 (Target)}\vphan}%
                \hspace{0.07in}%
                \stackon{\includegraphics[width=1.23in]{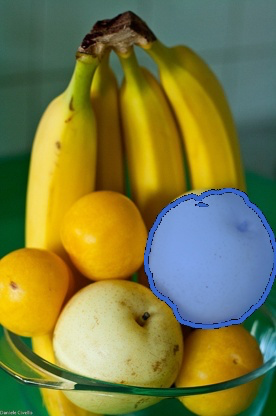}}{Grid 2\vphan}%
                \hspace{0.07in}%
                \stackon{\includegraphics[width=1.23in]{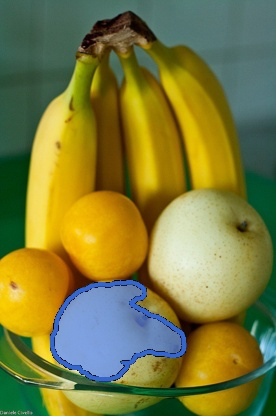}}{Grid 3\vphan}%
                \hspace{0.07in}%
                \stackon{\includegraphics[width=1.23in]{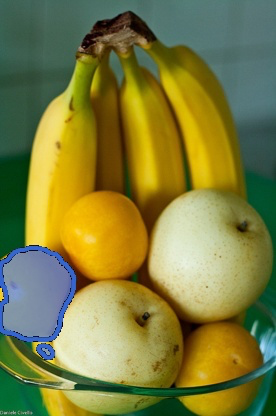}}{Grid 4\vphan}%
                \caption{\textit{Input: "banana in the back"}}
             \end{subfigure}%
          }%
 
          \vspace{0.1in}
 
          \fbox{
             \begin{subfigure}[t]{1.25in}%
                \stackon{\includegraphics[width=1.23in]{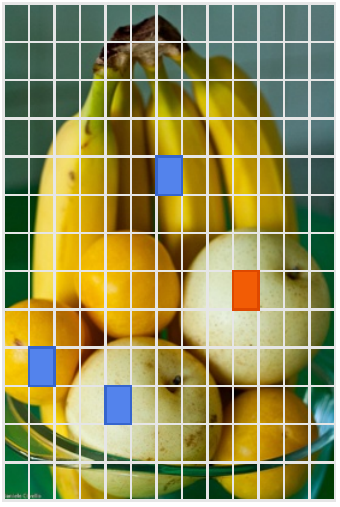}}{ \vphan}
                \caption{\textit{Image}}
             \end{subfigure}}%
          \hspace{0.1in}
          \fbox{
             \begin{subfigure}[t]{5.2in}
                \stackon{\includegraphics[width=1.23in]{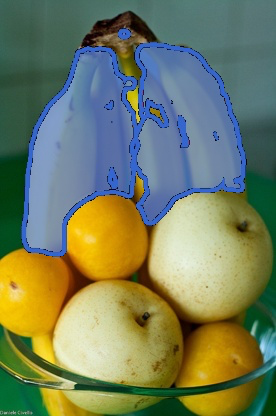}}{Grid 1\vphan}%
                \hspace{0.07in}%
                \stackon{\includegraphics[width=1.23in]{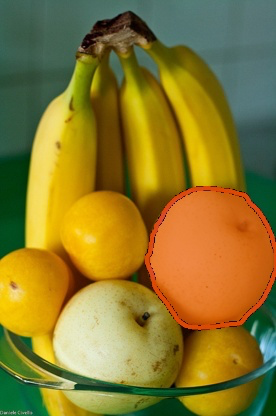}}{\textcolor{orange}{Grid 2 (Target)}\vphan}%
                \hspace{0.07in}%
                \stackon{\includegraphics[width=1.23in]{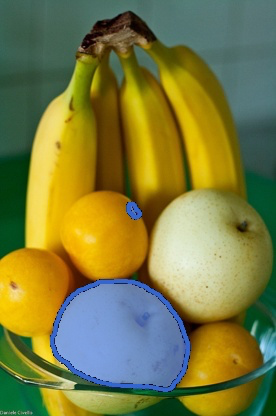}}{Grid 3\vphan}%
                \hspace{0.07in}%
                \stackon{\includegraphics[width=1.23in]{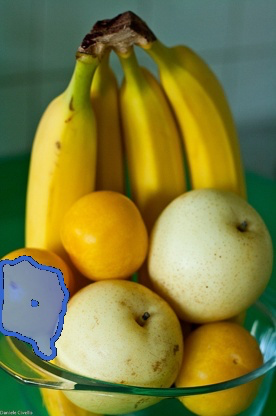}}{Grid 4\vphan}%
                \caption{\textit{Input: "pear on right"}}
             \end{subfigure}%
          }%
       \end{center}
       \vspace{0.05in}
       \caption{(Best viewed in color) Visualization of the output of the Segmentation Branch. The linguistic feature helps the Segmentation Branch to focus on generating fine-grained segmentation masks for instances that are more possible to be the target, but still aware of other instances.}
       \label{fig:eg_mask}
    \end{center}

\end{figure*}

%% file: bare_jrnl.bbl
\begin{thebibliography}{10}
\providecommand{\url}[1]{#1}
\csname url@samestyle\endcsname
\providecommand{\newblock}{\relax}
\providecommand{\bibinfo}[2]{#2}
\providecommand{\BIBentrySTDinterwordspacing}{\spaceskip=0pt\relax}
\providecommand{\BIBentryALTinterwordstretchfactor}{4}
\providecommand{\BIBentryALTinterwordspacing}{\spaceskip=\fontdimen2\font plus
\BIBentryALTinterwordstretchfactor\fontdimen3\font minus
  \fontdimen4\font\relax}
\providecommand{\BIBforeignlanguage}[2]{{%
\expandafter\ifx\csname l@#1\endcsname\relax
\typeout{** WARNING: IEEEtran.bst: No hyphenation pattern has been}%
\typeout{** loaded for the language `#1'. Using the pattern for}%
\typeout{** the default language instead.}%
\else
\language=\csname l@#1\endcsname
\fi
#2}}
\providecommand{\BIBdecl}{\relax}
\BIBdecl

\bibitem{he2016deep}
K.~He, X.~Zhang, S.~Ren, and J.~Sun, ``Deep residual learning for image
  recognition,'' in \emph{Proc. IEEE Conf. Comput. Vis. Pattern Recognit.},
  2016, pp. 770--778.

\bibitem{ding2020semantic}
H.~Ding, X.~Jiang, B.~Shuai, A.~Q. Liu, and G.~Wang, ``Semantic segmentation
  with context encoding and multi-path decoding,'' \emph{IEEE Trans. Image
  Processing}, vol.~29, pp. 3520--3533, 2020.

\bibitem{liu2021towards}
C.~Liu, H.~Ding, and X.~Jiang, ``Towards enhancing fine-grained details for
  image matting,'' in \emph{Proceedings of the IEEE/CVF Winter Conference on
  Applications of Computer Vision}, 2021, pp. 385--393.

\bibitem{ding2021interaction}
H.~Ding, H.~Zhang, J.~Liu, J.~Li, Z.~Feng, and X.~Jiang, ``Interaction via
  bi-directional graph of semantic region affinity for scene parsing,'' in
  \emph{Proceedings of the IEEE/CVF International Conference on Computer
  Vision}, 2021, pp. 15\,848--15\,858.

\bibitem{wang2021discovering}
S.~Wang, K.-H. Yap, H.~Ding, J.~Wu, J.~Yuan, and Y.-P. Tan, ``Discovering human
  interactions with large-vocabulary objects via query and multi-scale
  detection,'' in \emph{Proceedings of the IEEE/CVF International Conference on
  Computer Vision}, 2021, pp. 13\,475--13\,484.

\bibitem{ding2022deep}
H.~Ding, H.~Zhang, C.~Liu, and X.~Jiang, ``Deep interactive image matting with
  feature propagation,'' \emph{IEEE Transactions on Image Processing}, 2022.

\bibitem{shuai2018toward}
B.~Shuai, H.~Ding, T.~Liu, G.~Wang, and X.~Jiang, ``Toward achieving robust
  low-level and high-level scene parsing,'' \emph{IEEE Transactions on Image
  Processing}, vol.~28, no.~3, pp. 1378--1390, 2018.

\bibitem{liu2019feature}
J.~Liu, H.~Ding, A.~Shahroudy, L.-Y. Duan, X.~Jiang, G.~Wang, and A.~C. Kot,
  ``Feature boosting network for 3d pose estimation,'' \emph{IEEE transactions
  on pattern analysis and machine intelligence}, vol.~42, no.~2, pp. 494--501,
  2019.

\bibitem{chung2014empirical}
J.~Chung, C.~Gulcehre, K.~Cho, and Y.~Bengio, ``Empirical evaluation of gated
  recurrent neural networks on sequence modeling,'' \emph{arXiv preprint
  arXiv:1412.3555}, 2014.

\bibitem{hu2020dipnet}
P.~Hu, J.~Liu, G.~Wang, V.~Ablavsky, K.~Saenko, and S.~Sclaroff, ``Dipnet:
  Dynamic identity propagation network for video object segmentation,'' in
  \emph{Proc. IEEE Winter Conf. Appl. Comput. Vis.}, 2020, pp. 1904--1913.

\bibitem{hu2020uncertainty}
P.~Hu, S.~Sclaroff, and K.~Saenko, ``Uncertainty-aware learning for zero-shot
  semantic segmentation.'' in \emph{NeurIPS}, 2020.

\bibitem{zhang2021prototypical}
H.~Zhang and H.~Ding, ``Prototypical matching and open set rejection for
  zero-shot semantic segmentation,'' in \emph{Proc. IEEE Int. Conf. Comput.
  Vis.}, 2021, pp. 6974--6983.

\bibitem{kuen2019scaling}
J.~Kuen, F.~Perazzi, Z.~Lin, J.~Zhang, and Y.-P. Tan, ``Scaling object
  detection by transferring classification weights,'' in \emph{Proc. IEEE Int.
  Conf. Comput. Vis.}, 2019, pp. 6044--6053.

\bibitem{tong2021directed}
Z.~Tong, Y.~Liang, H.~Ding, Y.~Dai, X.~Li, and C.~Wang, ``Directed graph
  contrastive learning,'' \emph{Advances in Neural Information Processing
  Systems}, vol.~34, 2021.

\bibitem{ding2020phraseclick}
H.~Ding, S.~Cohen, B.~Price, and X.~Jiang, ``Phraseclick: toward achieving
  flexible interactive segmentation by phrase and click,'' in \emph{European
  Conference on Computer Vision}.\hskip 1em plus 0.5em minus 0.4em\relax
  Springer, 2020, pp. 417--435.

\bibitem{wu2020phrasecut}
C.~Wu, Z.~Lin, S.~Cohen, T.~Bui, and S.~Maji, ``Phrasecut: Language-based image
  segmentation in the wild,'' in \emph{Proc. IEEE Conf. Comput. Vis. Pattern
  Recognit.}, 2020, pp. 10\,216--10\,225.

\bibitem{ding2021vision}
H.~Ding, C.~Liu, S.~Wang, and X.~Jiang, ``Vision-language transformer and query
  generation for referring segmentation,'' in \emph{Proc. IEEE Int. Conf.
  Comput. Vis.}, 2021, pp. 16\,321--16\,330.

\bibitem{hu2016natural}
R.~Hu, H.~Xu, M.~Rohrbach, J.~Feng, K.~Saenko, and T.~Darrell, ``Natural
  language object retrieval,'' in \emph{Proc. IEEE Conf. Comput. Vis. Pattern
  Recognit.}, 2016, pp. 4555--4564.

\bibitem{hu2016segmentation}
R.~Hu, M.~Rohrbach, and T.~Darrell, ``Segmentation from natural language
  expressions,'' in \emph{Proc. Eur. Conf. Comput. Vis.}\hskip 1em plus 0.5em
  minus 0.4em\relax Springer, 2016, pp. 108--124.

\bibitem{luo2020multi}
G.~Luo, Y.~Zhou, X.~Sun, L.~Cao, C.~Wu, C.~Deng, and R.~Ji, ``Multi-task
  collaborative network for joint referring expression comprehension and
  segmentation,'' in \emph{Proc. IEEE Conf. Comput. Vis. Pattern Recognit.},
  2020, pp. 10\,034--10\,043.

\bibitem{ye2019cross}
L.~Ye, M.~Rochan, Z.~Liu, and Y.~Wang, ``Cross-modal self-attention network for
  referring image segmentation,'' in \emph{Proc. IEEE Conf. Comput. Vis.
  Pattern Recognit.}, 2019, pp. 10\,502--10\,511.

\bibitem{yu2018mattnet}
L.~Yu, Z.~Lin, X.~Shen, J.~Yang, X.~Lu, M.~Bansal, and T.~L. Berg, ``Mattnet:
  Modular attention network for referring expression comprehension,'' in
  \emph{Proc. IEEE Conf. Comput. Vis. Pattern Recognit.}, 2018, pp. 1307--1315.

\bibitem{he2017mask}
K.~He, G.~Gkioxari, P.~Doll{\'a}r, and R.~Girshick, ``Mask r-cnn,'' in
  \emph{Proc. IEEE Int. Conf. Comput. Vis.}, 2017, pp. 2961--2969.

\bibitem{yolo}
J.~Redmon, S.~Divvala, R.~Girshick, and A.~Farhadi, ``You only look once:
  Unified, real-time object detection,'' in \emph{Proc. IEEE Conf. Comput. Vis.
  Pattern Recognit.}, 2016, pp. 779--788.

\bibitem{wang2019solo}
X.~Wang, T.~Kong, C.~Shen, Y.~Jiang, and L.~Li, ``Solo: Segmenting objects by
  locations,'' in \emph{Proc. Eur. Conf. Comput. Vis.}\hskip 1em plus 0.5em
  minus 0.4em\relax Springer, 2020, pp. 649--665.

\bibitem{ye2020dual}
L.~Ye, Z.~Liu, and Y.~Wang, ``Dual convolutional lstm network for referring
  image segmentation,'' \emph{IEEE Transactions on Multimedia}, 2020.

\bibitem{chen2019referring}
Y.-W. Chen, Y.-H. Tsai, T.~Wang, Y.-Y. Lin, and M.-H. Yang, ``Referring
  expression object segmentation with caption-aware consistency,''
  \emph{Linguistic Structure Guided Context Modeling for Referring Image
  Segmentation}, 2019.

\bibitem{long2015fully}
J.~Long, E.~Shelhamer, and T.~Darrell, ``Fully convolutional networks for
  semantic segmentation,'' in \emph{Proc. IEEE Conf. Comput. Vis. Pattern
  Recognit.}, 2015, pp. 3431--3440.

\bibitem{ding2019boundary}
H.~Ding, X.~Jiang, A.~Q. Liu, N.~M. Thalmann, and G.~Wang, ``Boundary-aware
  feature propagation for scene segmentation,'' in \emph{Proc. IEEE Int. Conf.
  Comput. Vis.}, 2019, pp. 6819--6829.

\bibitem{ding2019semantic}
H.~Ding, X.~Jiang, B.~Shuai, A.~Q. Liu, and G.~Wang, ``Semantic correlation
  promoted shape-variant context for segmentation,'' in \emph{Proc. IEEE Conf.
  Comput. Vis. Pattern Recognit.}, 2019, pp. 8885--8894.

\bibitem{liu2017recurrent}
C.~Liu, Z.~Lin, X.~Shen, J.~Yang, X.~Lu, and A.~Yuille, ``Recurrent multimodal
  interaction for referring image segmentation,'' in \emph{Proc. IEEE Int.
  Conf. Comput. Vis.}, 2017, pp. 1271--1280.

\bibitem{shi2018key}
H.~Shi, H.~Li, F.~Meng, and Q.~Wu, ``Key-word-aware network for referring
  expression image segmentation,'' in \emph{Proc. Eur. Conf. Comput. Vis.},
  2018, pp. 38--54.

\bibitem{luo2017comprehension}
R.~Luo and G.~Shakhnarovich, ``Comprehension-guided referring expressions,'' in
  \emph{Proc. IEEE Conf. Comput. Vis. Pattern Recognit.}, 2017, pp. 7102--7111.

\bibitem{yu2017joint}
L.~Yu, H.~Tan, M.~Bansal, and T.~L. Berg, ``A joint speaker-listener-reinforcer
  model for referring expressions,'' in \emph{Proc. IEEE Conf. Comput. Vis.
  Pattern Recognit.}, 2017, pp. 7282--7290.

\bibitem{zhang2017discriminative}
Y.~Zhang, L.~Yuan, Y.~Guo, Z.~He, I.-A. Huang, and H.~Lee, ``Discriminative
  bimodal networks for visual localization and detection with natural language
  queries,'' in \emph{Proc. IEEE Conf. Comput. Vis. Pattern Recognit.}, 2017,
  pp. 557--566.

\bibitem{wang2019neighbourhood}
P.~Wang, Q.~Wu, J.~Cao, C.~Shen, L.~Gao, and A.~v.~d. Hengel, ``Neighbourhood
  watch: Referring expression comprehension via language-guided graph attention
  networks,'' in \emph{Proc. IEEE Conf. Comput. Vis. Pattern Recognit.}, 2019,
  pp. 1960--1968.

\bibitem{yang2019dynamic}
S.~Yang, G.~Li, and Y.~Yu, ``Dynamic graph attention for referring expression
  comprehension,'' in \emph{Proc. IEEE Int. Conf. Comput. Vis.}, 2019, pp.
  4644--4653.

\bibitem{lin2017feature}
T.-Y. Lin, P.~Doll{\'a}r, R.~Girshick, K.~He, B.~Hariharan, and S.~Belongie,
  ``Feature pyramid networks for object detection,'' in \emph{Proc. IEEE Conf.
  Comput. Vis. Pattern Recognit.}, 2017, pp. 2117--2125.

\bibitem{ding2018context}
H.~Ding, X.~Jiang, B.~Shuai, A.~Q. Liu, and G.~Wang, ``Context contrasted
  feature and gated multi-scale aggregation for scene segmentation,'' in
  \emph{Proc. IEEE Conf. Comput. Vis. Pattern Recognit.}, 2018, pp. 2393--2402.

\bibitem{pennington2014glove}
J.~Pennington, R.~Socher, and C.~D. Manning, ``Glove: Global vectors for word
  representation,'' in \emph{Proc. Conf. Empir. Methods Nat. Lang. Process.},
  2014, pp. 1532--1543.

\bibitem{chen2017deeplab}
L.-C. Chen, G.~Papandreou, I.~Kokkinos, K.~Murphy, and A.~L. Yuille, ``Deeplab:
  Semantic image segmentation with deep convolutional nets, atrous convolution,
  and fully connected crfs,'' \emph{IEEE Trans. Pattern Anal. Mach. Intell.},
  vol.~40, no.~4, pp. 834--848, 2017.

\bibitem{margffoy2018dynamic}
E.~Margffoy-Tuay, J.~C. P{\'e}rez, E.~Botero, and P.~Arbel{\'a}ez, ``Dynamic
  multimodal instance segmentation guided by natural language queries,'' in
  \emph{Proceedings of the European Conference on Computer Vision (ECCV)},
  2018, pp. 630--645.

\bibitem{li2018referring}
R.~Li, K.~Li, Y.-C. Kuo, M.~Shu, X.~Qi, X.~Shen, and J.~Jia, ``Referring image
  segmentation via recurrent refinement networks,'' in \emph{Proc. IEEE Conf.
  Comput. Vis. Pattern Recognit.}, 2018, pp. 5745--5753.

\bibitem{huang2020referring}
S.~Huang, T.~Hui, S.~Liu, G.~Li, Y.~Wei, J.~Han, L.~Liu, and B.~Li, ``Referring
  image segmentation via cross-modal progressive comprehension,'' in
  \emph{Proc. IEEE Conf. Comput. Vis. Pattern Recognit.}, 2020, pp.
  10\,488--10\,497.

\bibitem{hui2020linguistic}
T.~Hui, S.~Liu, S.~Huang, G.~Li, S.~Yu, F.~Zhang, and J.~Han, ``Linguistic
  structure guided context modeling for referring image segmentation,''
  \emph{Proc. Eur. Conf. Comput. Vis.}, 2020.

\bibitem{luo2020cascade}
G.~Luo, Y.~Zhou, R.~Ji, X.~Sun, J.~Su, C.-W. Lin, and Q.~Tian, ``Cascade
  grouped attention network for referring expression segmentation,'' in
  \emph{Proceedings of the 28th ACM International Conference on Multimedia},
  2020, pp. 1274--1282.

\bibitem{chen2019see}
D.-J. Chen, S.~Jia, Y.-C. Lo, H.-T. Chen, and T.-L. Liu, ``See-through-text
  grouping for referring image segmentation,'' in \emph{Proc. IEEE Int. Conf.
  Comput. Vis.}, 2019, pp. 7454--7463.

\bibitem{hu2020bi}
Z.~Hu, G.~Feng, J.~Sun, L.~Zhang, and H.~Lu, ``Bi-directional relationship
  inferring network for referring image segmentation,'' in \emph{Proc. IEEE
  Conf. Comput. Vis. Pattern Recognit.}, 2020, pp. 4424--4433.

\bibitem{yu2016modeling}
L.~Yu, P.~Poirson, S.~Yang, A.~C. Berg, and T.~L. Berg, ``Modeling context in
  referring expressions,'' in \emph{Proc. Eur. Conf. Comput. Vis.}\hskip 1em
  plus 0.5em minus 0.4em\relax Springer, 2016, pp. 69--85.

\bibitem{mao2016generation}
J.~Mao, J.~Huang, A.~Toshev, O.~Camburu, A.~L. Yuille, and K.~Murphy,
  ``Generation and comprehension of unambiguous object descriptions,'' in
  \emph{Proc. IEEE Conf. Comput. Vis. Pattern Recognit.}, 2016, pp. 11--20.

\bibitem{nagaraja2016modeling}
V.~K. Nagaraja, V.~I. Morariu, and L.~S. Davis, ``Modeling context between
  objects for referring expression understanding,'' in \emph{Proc. Eur. Conf.
  Comput. Vis.}\hskip 1em plus 0.5em minus 0.4em\relax Springer, 2016, pp.
  792--807.

\end{thebibliography}
